# Towards Multi-Modal Mastery: A 4.5B Parameter Truly Multi-Modal Small Language Model

Ben Koska
*Informatics*
TU Wien
Vienna, Austria
ben.koska@student.tuwien.ac.at

Mojmír Horváth
*Informatics*
TU Wien
Vienna, Austria
mojmir.horvath@student.tuwien.ac.at

*Abstract*—We present a novel 4.5B parameter small language model that can handle multiple input and output modalities, including text, images, videos, and audio. Despite its small size, the model achieves near state-of-the-art performance on a variety of tasks, demonstrating the potential of multi-modal models to tackle complex real-world problems. Our approach leverages recent advancements in language modeling and multi-task learning to create a versatile and high-performing model that can even be deployed for edge inference. Experimental results show the model's strong performance across multiple benchmarks, paving the way for further progress in multi-modal artificial intelligence.

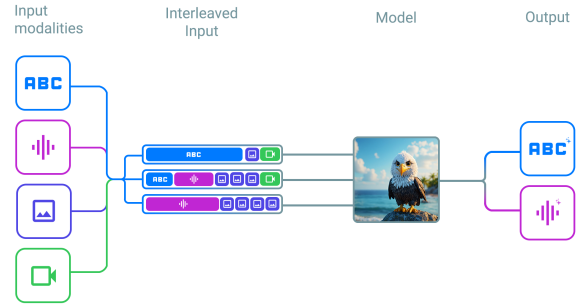

Fig. 1 EAGLE Model overview

## I. Introduction

Humans interact with the world through multiple senses - sight, sound, touch, and language, each providing complementary information that helps us understand and reason about our environment. One of the primary goals in Artificial Intelligence has been to develop a general-purpose assistant that can mimic this multi-modal intelligence, processing and generating a diverse range of information [1]. However, current large language models (LLMs) still struggle to effectively handle non-textual inputs and outputs, limiting their applicability to real-world scenarios.

Recent works focusing on providing LLMs with a second sense in addition to text, namely vision, have shown promising results [2] [3] [4]. Recent works further demonstrate improved performance through techniques such as increasing the pre-training data [5] or scaling up the vision encoder [6]. To evaluate the performance of Multi-Modal LLMs, various benchmarks have been proposed [7] [8] [9] [10].

Furthermore, many models focus on text-image pairs [2] [5] [6] or more recently text-video pairs [11] [12]. To allow for a natural way of interacting with a general assistant, there exists a need to interact using natural speech instead of just using text. Recent works on allowing more modalities than just text and visuals show promising results [13] [14] [3].

While some works explore the usage of small models [15], most still utilize models too large to run on-edge (on a smart phone or laptop) causing the need for expensive hardware for inference, a stable internet connection on devices and provide an attack surface for malicious actors.

However, despite the promising results from recent research on multi-modal language models, no standard recipe has been established for training these models effectively. Different techniques have shown improvements on various benchmarks, but their performance can vary widely across different tasks and datasets. This lack of a unified approach highlights the need for further exploration and experimentation to develop a more robust and generalized multi-modal modeling framework.

To this extent, in this paper we present EAGLE (4.3B parameters), a large-language model with vision, audio and text input capabilities which outputs text, as well as EAGLE-Assistant (4.5B parameters), which extends the capabilities of EAGLE to allow it to output audio, enabling an end-to-end (audio-in, audio-out), natural verbal conversation between the user and model.

## II. Approach

### A. Architecture

The architecture of EAGLE (see Fig. 3) combines 3 components, a large-language model (phi-3-mini [16], 128K context window variant - 3.7B parameters), an audio tower (whisper-small [17] - 244M parameters) and a vision tower (CLIP [18] ViT Large Patch14 - 303M parameters).

| Modality | Tokens per Sample |
|---|---|
| Image | $\lceil \frac{height}{336} \rceil \times \lceil \frac{width}{336} \rceil \times 128$ |
| Speech | ~3 tokens per second[1] |

Fig. 2. Tokens per Sample for different modalities

As the output space of the audio tower, vision tower and language model differ we employ a projection layer. The projection layers project the audio tower and vision tower respectively into the output space of the language model. The projected tokens are then combined in an interleaved manner (no particular order). Following [16], we utilize a dynamic cropping strategy [19] to accommodate dynamic-resolution and various aspect ratios. This is achieved by splitting the images into a 2D array of blocks (336px by 336px resolution) which are then flattened to represent the entire image. The required tokens per image therefore differ depending on image resolution (see Fig. 2).

---

[1] Due to the tokenization exact tokens per second varies widely, we observe on average 2.7 tokens per second across a wide array of German, English, and Spanish texts. For Slovak we observe on average 3.4 tokens per second.

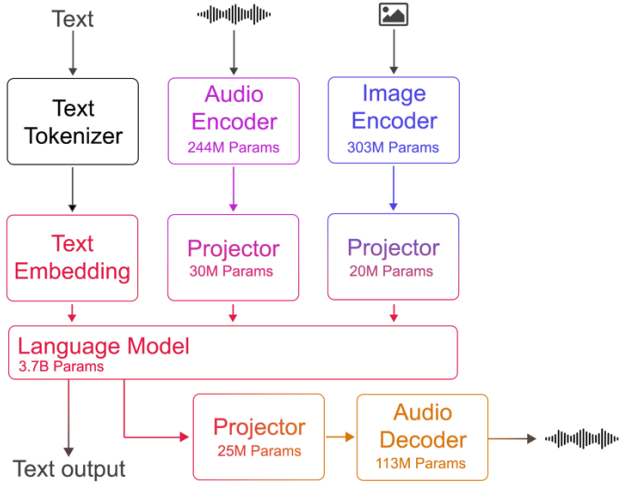

Fig. 3. Architecture Diagram

The model requires at least a text or audio input to function properly. While we experimented with image-only inputs, we did not manage to obtain sufficient training data to produce a valuable result, which we leave to future works.

For EAGLE-A we add another module, based on the architecture of OpenVoice [20] for audio output. EAGLE-A is further finetuned for chat and function-calling support.

### B. Training

We initialize the language model using the weights from Phi3.5 mini long context (128K tokens).

**Pre-training**

For pre-training we utilize a two-stage approach.

<u>Stage 1: Pre-training projection.</u> In this stage we freeze the image encoder, audio encoder and language model. The projectors are randomly initialized and are then trained on a random 30M token subset of our pre-training dataset.

<u>Stage 2: Full-parameter fine-tuning:</u> In this stage we unfreeze all modules and train on the remainder of the pre-training dataset.

**Fine-tuning:**

For training of the audio decoder and all subsequent fine-tuning (Instruction tuning, chat-tuning and Function-calling) we keep all modules unfrozen.

### C. Data

For pre-training, we utilize a custom dataset which consists of a combination of image-text pair datasets (e.g., LAION-COCO), interleaved image-text document datasets (e.g., OBELICS [21]), synthetic OCR Data (e.g. RenderedText) and real-world OCR Data (e.g. IDL [22], PDFA), a synthetic audio version of TriviaQA [23], speech data (e.g. LibriSpeech [24]) and self-generated synthetic data. For better understanding of emotion, non-text verbal clues (e.g. such as coughing, sarcastic voice) and speed/volume of voice we a) train a transcription model with tokens to identify theseand transcribe thousands of hours of speech with said model and b) generate synthetic speech using Text-to-Speech (TTS) and Speech-to-Speech (STS) models.

For synthetic data we utilize a combination of approaches:

*1) **Real-base-synthetic-data:** We utilize real data, such as PDFs, Charts and Images which we then use as a direct base to generate synthetic data (such as generating a converstaion about a specific PDF or turning a text in a conversation into audio using text-to-speech)*

*2) **Double-Synthetic:** To cover cases which lack a substantial amount of (accessible) real base data, we utilize a double-synthetic approach. In this approach we generate synthetic charts, letters, images, etc. which are used either as a base to create synthetic conversations, or are created in conjunction with the synthetic data (e.g. conversations).*

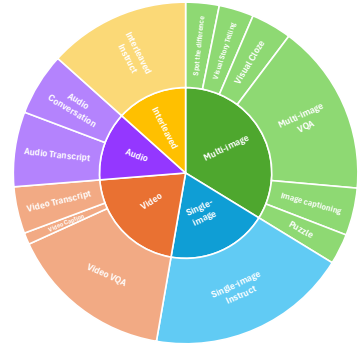

Fig. 4. Data Statistic of instruction-tuning dataset

For instruction-tuning, we utilize *The Cauldron* [25] collection (A collection of 50 vision-language training datasets – 462M tokens + 3.7M images) as well as a custom dataset made up of a variety of multi-modal data (see Fig. 4), sourced from various datasets with minor synthetic additions.

For chat-tuning, we utilize AnyInstruct [13] and a custom dataset of 90k conversation, containing 45k synthetic conversations, 15k genuine multi-modal LLM conversations (real users talking with Claude or GPT-4o/GPT-4v) sourced from a proprietary dataset and 30k "semi-synthetic" conversations created by taking the genuine conversations and either a) translating them into another language or b) rephrasing them.

For function-calling, we utilize all 60k rows of the APIGen Function-Calling Datasets [26], a subset of 50k rows of the dataset *glaiveai/glaive-function-calling-v2 (*wherein multi-turn conversations were prioritized in subset selection), as well as a dataset of 25k (20k single-turn, 5k multi-turn) synthetic rows of multi-modal function-calling created by interleaving text, speech (text-to-speech + random occasional background noise) and images.

| | | Open-Source | | | Proprietary | | | | |
|---|---|---|---|---|---|---|---|---|---|
| | EAGLE | Phi-3-vision | LLAVA-NeXT | InternVL2 | Gemini 1.5 Pro | GPT-4o | Claude 3.5 | SoTA | Human Expert |
| **Parameters** | 4.3B | 4.2B | 34B | 40B | — (100B+) | — (100B+) | — | | — |
| **Context Window** | 128k | 128k | 4k | 8k | 128k | 128k | 200k | | |
| **$ / million tokens** | $0.08 | — | — | — | $10.50 | $15.00 | $15.00 | — | $$$ |
| **MMMU** ↑ | 46.3 | 40.4 | 51.1 | 53.9 | 62.2 | **69.1** | 68.3 | 69.1 | 76.2 |
| **ScienceQA** ↑ | **94.6** | 90.8 | 81.8 | — | — | 83.9 | — | 96.1 | 88.4 |
| **MMLU** ↑ | 72.9 | 68.1 | — | — | 78.50 | **88.7** | 88.3 | 90.0 | 89.8 |
| **ChartQA** ↑ | 84.4 | 81.4 | 68.7 | 86.2 | 81.3 | 85.7 | **90.8** | 90.8 | — |
| **MMBench** ↑ | 80.1 | 73.6 | 81.1 | **86.8** | 73.9 | 83.4 | 79.7 | 85.5 | — |
| **Audio ASR** ↓ | 02.6 | — | | | — | | | 01.4 | 05.8 |
| **AudioCaps** ↑ | 86.3 | | | | | | | 83.2 | 91.3 |

Table 1. Academic Benchmark Evaluation

For speech output, we utilize our custom-labeled dataset from pre-training. We utilize 7 datasets to train 7 distinct voices (Chris, Mia, Jim, Emma, Tom, Lucy, and Alex). The datasets of 5 voices are created synthetically using a commercial text-to-speech tool and the datasets for 2 voices are proprietary.

### III. EVALUATION

In Table 1 we report the results for EAGLE on standard open-source benchmarks measuring the model's reasoning, vision and audio ability. We compare EAGLE to Phi-3-vision [16], LLAVA-NeXT-34B [27], InternVL2-40B [28], Gemini 1.5 Pro [29], GPT-4o [30] and Claude 3.5 Sonnet [31]. The table is a summation of publicly published numbers. For benchmarks with public leaderboard (e.g. MMBench [9] and MMMU) preference is given to the results on published the leaderboard. EAGLE is evaluated in the manner that is standard for each benchmark and an effort is made to ensure that all values for other models follow the same method of evaluation. Due to a lack of accessible modern LLMs with audio capabilities, which also report other relevant benchmarks such as MMMU and MMBench EAGLE stands alone as an LLM in the category of audio benchmarks. For price per million tokens, we report output prices and, in the case of Gemini 1.5 Pro we utilize the base 128k context window model.

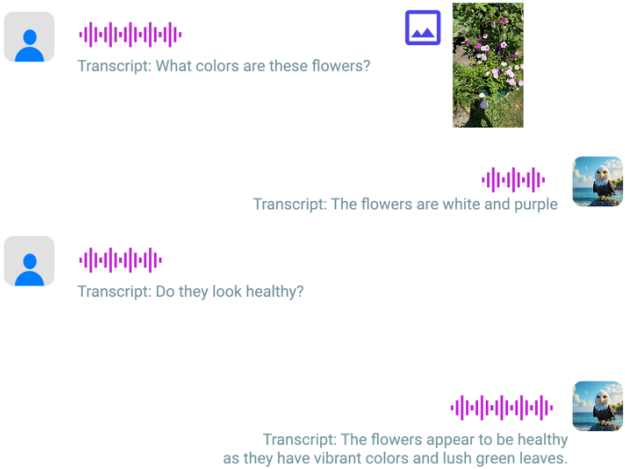

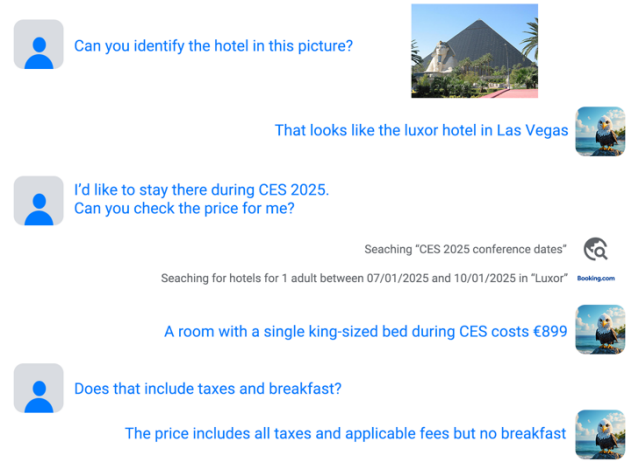

Fig. 5. Example of assistant usage of EAGLE-A in end-to-end audio mode. System prompt is provided as text, all assistant and user questions are provided directly as audio with transcription.

Fig. 6. Example of assistant usage of EAGLE-A with function calling to identify a hotel from a picture and then check room prices for the duration of CES 2025.

We evaluate function calling accuracy using our internal assistant function calling benchmark (see example in Fig. 6). We test function calling with a) functions provided in the system prompt, as well as b) fine-tuning EAGLE-A with a small (750 samples per function) synthetic dataset of calling the available functions, without providing them in the system prompt. We do not notice a meaningful difference between one method and the other (97.2% in-context vs 97% fine-tuned) but do observe a significant drop (from 97% to 89.5%) in accuracy using functions provided through fine-tuning in cases where fine-tuning and in-context are mixed.

As modern LLMs with voice output emerge [30] [13] [32], the necessity for voice-output LLM benchmarks becomes apparent. While we have developed an internal benchmark to evaluate training progress and conduct experiments, it is not sufficiently robust to comprehensively assess a wide array of models. Consequently, the creation of a comprehensive and robust benchmark is left to future work.

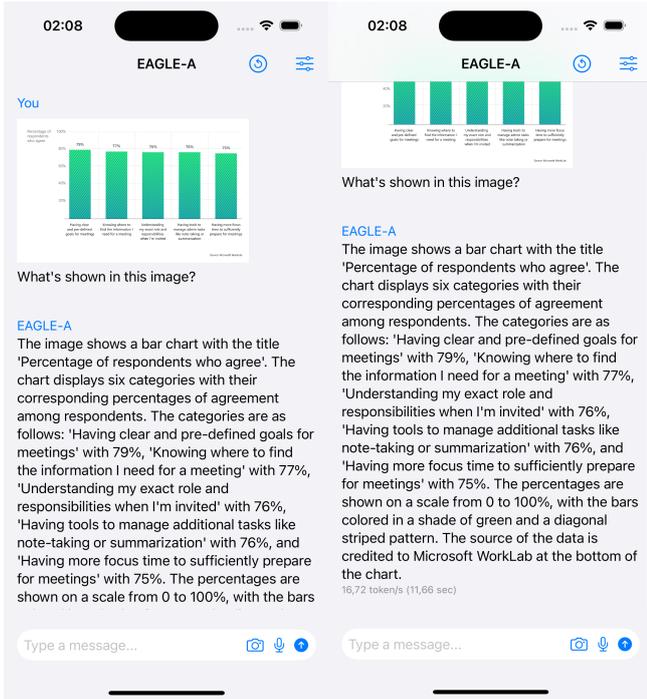
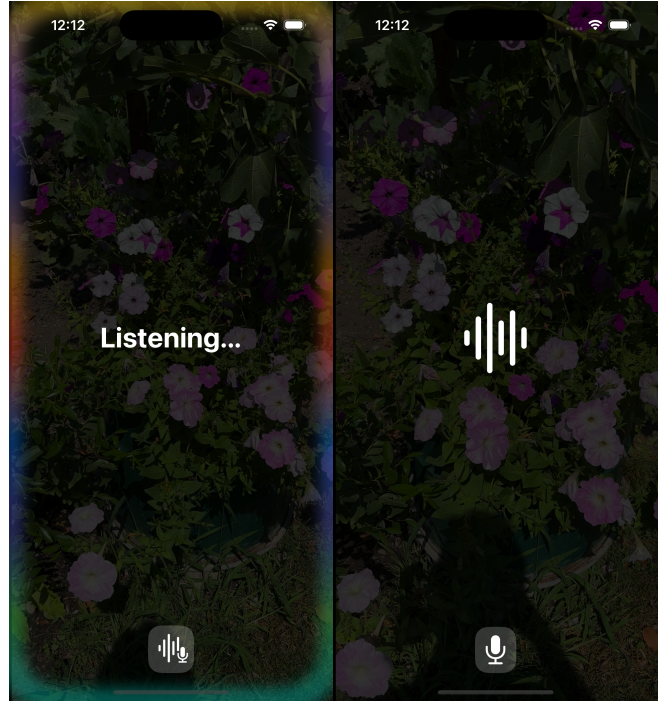

| Text + Image chat | Assistant with audio-in, audio-out + image modality via camera |

Fig. 7. **EAGLE-A** running natively on an iPhone 15 Pro with A17 Pro with a performance of more than 16 tokens per second.

## IV. EDGE INFERENCE

To efficiently deploy the model on-edge we utilize several strategies:
1) Mixed-precision quantization
2) Hand-optimized implementation
3) Quantization-aware full-parameter fine-tuning

Running the final version of the mobile model on an iPhone 15 Pro (see Fig. 7) yields a generation speed of nearly 17 tokens per second. Furthermore, running the EAGLE-A model in full end-to-end (audio-in, audio-out) voice assistant model is supported, and yields above real-time generation and a time-to-first-token (TTFT) of 425ms, allowing for natural, real-time voice communication with the assistant.

### A. Quantization-Aware Fine-Tuning

Building upon the work of [33] we developed a new training technique, which we then utilized to fine-tune the base model at different quantization settings. By utilizing full-parameter quantization-aware fine-tuning we manage to regain most of the performance loss of quantization (see Table 2). Using our mixed-precision quantization configuration (resulting in, on average 5.5-bits per parameter), we manage to reduce the model size from 18GB (at float32) to just over 3GB, allowing the model to fit into the memory of most modern smart phones. Our experiments show that full parameter tuning yields significantly better results.

| Data Type | Fine-tuning | Result (%) ↑ |
|---|---|---|
| float32 | - | 75.3% |
| bfloat16 | - | 75.4% |
| int8 | - | 75.2% |
| | w/ quantization-aware full-parameter fine-tuning | 75.4% |
| | w/ quantization-aware LoRA | 75.2% |
| int4 | - | 70.3% |
| | w/ quantization-aware full-parameter fine-tuning | 73.8% |
| | w/ quantization-aware LoRA | 72.6% |
| Mixed-precision (5.5-bits) | - | 71.9% |
| | w/ quantization-aware full-parameter fine-tuning | 75.1% |
| | w/ quantization-aware LoRA | 72.6% |

Table 2. Result of quantization, and quanization-aware fine-tuning on benchmarks. We report average scores across a our internal evaluation suite.

## V. MODEL CARD

| Metric | |
|---|---|
| Training Time | 39 hours |
| GPUs | 32 nodes of 8x NVIDIA H100 (256 H100s) |
| Training Date | September 2024 |

Table 3. Model card

## VI. Conclusion

We introduced EAGLE and EAGLE-A, two compact multi-modal models with 4.3 and 4.5 billion parameters, capable of processing and generating text, images, audio, and video. Despite their smaller size, both models achieve competitive performance across various benchmarks, showcases new possibilities of on edge computing and new ways to think about large language models.

Key innovations in these models include the integration of multiple modality towers and quantization-aware fine-tuning, enabling efficient deployment on resource-constrained devices. While the models perform well, limitations such as the need for more robust image-only training data and reliance on synthetic datasets remain.

Future work will focus on improving training efficiency, expanding the model's capabilities to handle broader tasks, and developing better multi-modal benchmarks, particularly for audio-based tasks. EAGLE represents a step toward more capable and versatile multi-modal LLM system, paving the way for advanced general-purpose assistants without the need of a heavy-duty GPU.